\documentclass[10pt,twocolumn,letterpaper]{article}

\usepackage{cvpr}
\usepackage{times}
\usepackage{epsfig}
\usepackage{graphicx}
\usepackage{amsmath}
\usepackage{amssymb}
\usepackage{authblk}
\cvprfinalcopy % *** Uncomment this line for the final submission

\begin{document}

%%%%%%%%% TITLE
\title{ParallelNet: Multi-mode Trajectory Prediction by Multi-mode Trajectory Fusion}

%\author{Luoyu Chen, Hao Lu, Fei Wu\\
%Australian National University\\
%Canberra ACT 2601\\
%{\tt\small first.second@anu.edu.au}
% \and
%\author[1]{Author %A\thanks{A.A@university.edu}}
\author[1]{Luoyu Che{n}$^\dag$$^*$}
%\author[1]{LuoYu Che{n}}
\author[1]{Fei W{u}$^\dag$$^\ddagger$}
%\author[1]{Fei W{u}}
%\author[1,2]{Hao L{u}$^\dag$}
\author[1,2]{Hao L{u}$^\dag$}

\affil[1]{Department of Computer Engineering, Australian National University}
\affil[2]{NXP® Semiconductors }
\twocolumn[{%
\renewcommand\twocolumn[1][]{#1}%
\maketitle
\begin{center}
    \centering
    \includegraphics[width=1\textwidth,height=5cm]{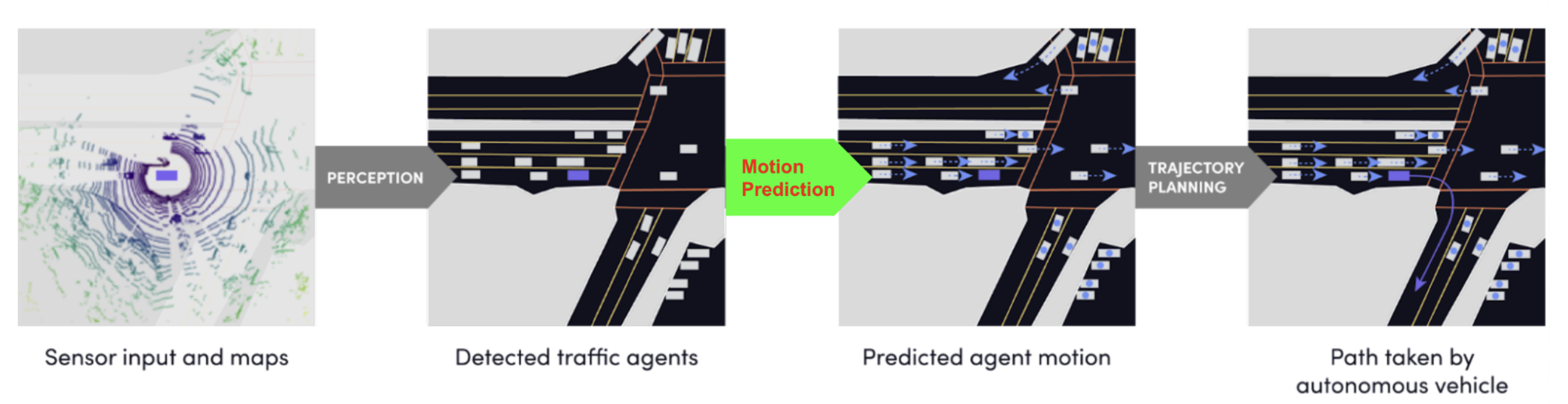}
\end{center}%
}]

%\maketitle
\def\thefootnote{$\dag$}\footnotetext{These authors contributed equally to this work}
\def\thefootnote{$*$}\footnotetext{luoyu.chen.mlcv@gmail.com}
\def\thefootnote{$\ddagger$}\footnotetext{wufei.mlcv@gmail.com}
%\thispagestyle{empty}

%%%%%%%%% ABSTRACT

\begin{abstract}
Level 5 Autonomous Driving, a technology that a fully automated vehicle (AV) requires no human intervention, has raised serious concerns on safety and stability before widespread use. The capability of understanding and pre- dicting future motion trajectory of road objects can help AV plan a path that is safe and easy to control. In this paper, we propose a network architecture that parallelizes multiple convolutional neural network backbones and fuses features to make multi-mode trajectory prediction. In the 2020 ICRA Nuscene Prediction challenge, our model ranks 15th on the leaderboard across all teams.
\end{abstract}

%%%%%%%%% BODY TEXT

\section{Introduction}
 In general, the whole pipline as shown in open image(https://level-5.global/data/), given a 3D cloud points by sensor to perceive traffic agents, then doing motion prediction and trajectory planning. 
One of the key components in autonomous driving (AV) is motion prediction. Trajectory predictions for on-road agents such as cars, cyclists, pedestrians can help AV have a better understanding of future actions of surrounding agents so that adequate information can be provided to AV decision-making module to make safe and user comfortable actuation planning. However, motion prediction is still a challenging task, since a robust motion prediction system has to adapt to any arbitrary landscape and on-road event. This paper is using one of the most popular image-based methods: using rasterized Bird Eye View (BEV) map and kinematic data to predict multiple possible future trajecto- ries with confidences in the next 6 seconds. Our model is trained and evaluated on the Nuscene prediction challenge dataset and ranks 15th place on the leaderboard for overall criterion, 7th on the off-road rate. However, to achieve real- time inference speed, our model is trained on the exact mo- ment data instead of using any history data, which includes road scene and every on-road agent kinematic data in the past 2 seconds. Using 2 seconds history data is allowed in this challenge, so our model can be more competitive if we take account of history data and choose to use more compu- tational resources.
\begin{figure*}[t]
\begin{center}

\includegraphics[width=\textwidth]{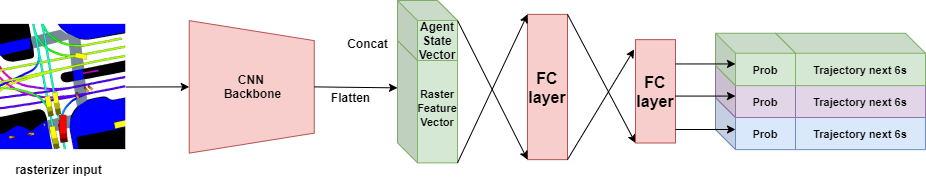}
\end{center}
  \caption{The architecture of MTP model}
\label{fig:long}
\label{fig:onecol}
\end{figure*}
%-------------------------------------------------------------------------

\section{Related Research}
The easiest and most popular feature extraction is sim- ply feeding rasterized BEV map into the CNN backbone, however, rasterized BEV map consists of multiple static layers describing the landscape and dynamic layers describ- ing agents motion, which differs at every moment. There- fore, if rasterize the complete BEV map, dynamic layers can cover areas where static layers have been rasterized, caus- ing information loss. However, if separate as different lay- ers will add the number of channels, significantly increase inference time and too expensive for real-time prediction. VectorNet ~\cite{gao2020vectornet} perform spatial-temporal discrete sampling to obtain points on a map, instead of using CNN kernels to ex- tract semantic information. Embedded sampled point with valid location, feature, Id information as a vertex of a typical sub-graph, and all identified sub-graphs form a global inter- action graph as a graph neural network, finally use MLP as decoder to predict multiple trajectories. This method over- comes CNN’s local attention shortcomings, which can not attend long-distance objects such as lanes effectively. Also, the graph embedding significantly reduced time complex- ity compared to traditional CNN scene parsing. Similarly, instead of using the original HD map directly, Ming et el. proposed LaneGCN ~\cite{liang2020learning}, HD map is formed as Lane graph to keep map structure and extract complex long-distance topological relationship and dependency. In Lane graph, each lane is a sequence of nodes, two lanes are reachable in between themselves conditioning on no presence of traffic rule violation, an adjacent matrix is used to describe four basic relationships between two lanes: predecessor, succes- sor, left, and right. This structural encoding of HD map lanes provides basic geometric and semantic relations to constrain predicted trajectory, drastically reducing the off lane rate. Our method is using multiple CNN backbones to extract rasterized HD maps, since it is simple for imple- mentation, and multiple CNN backbones can synchronize the feature extraction on different rasterized layers if using multiple GPU to speed up.\\

 \textbf{Traffic and road interaction modelling as complex system:}
 Vehicle future trajectories are generally coupled with road landscape, so possible future trajectories can be treated as a conditional probability distribution. Hang et el. proposed TNT ~\cite{zhao2021tnt}, treat multi-mode trajectory prediction as a two-phase serial decision making problem: target proposal, target prediction, motion estimation, trajectory scoring. In the first stage, multiple destinations are being proposed conditioning on the road landscape uncertainty and human actuation uncertainty, and in the second stage, by taking account of these uncertainties some reasonable targets are selected as prediction phase. Incorporating this uncertainty can add more diversity in predicted trajectories and be highly flexible in arbitrary road and traffic scenarios. Cui et el. proposed LookOut ~\cite{cui2021lookout}, by modelling multi-agent future trajectories joint distribution, a diverse set of future scenes can be estimated, so that vehicle trajectory prediction can be constructed by optimizing contingency plans for all possible future realization, safe and non-conservative trajectories will be planned base on that. Scene Transformer~\cite{ngiam2021scene} use MLP to generate time step features for the entire set of agents, and use PointNet~\cite{qi2017pointnet} to generate vectorized static road representation. Then use attention block to encode the temporal features of agents and spatial features of the road. Finally use attention decoder to output the joint distribution of multi-agent trajectory, achieving better conformity than marginalized single-agent trajectory prediction. Our work did not explicitly model the dynamic interactive environment for on-road agents and road landscape, we embedded that on feature extraction phase, by rasterizing multiple types of composite layers consisting of lane with agents, drivable area with agents, ego agents with other agents into CNN backbone to automatically learn some latent relations in between these composite layers.\\
 \begin{figure*}[h!]
\begin{center}
\includegraphics[width=\textwidth, height=10cm]{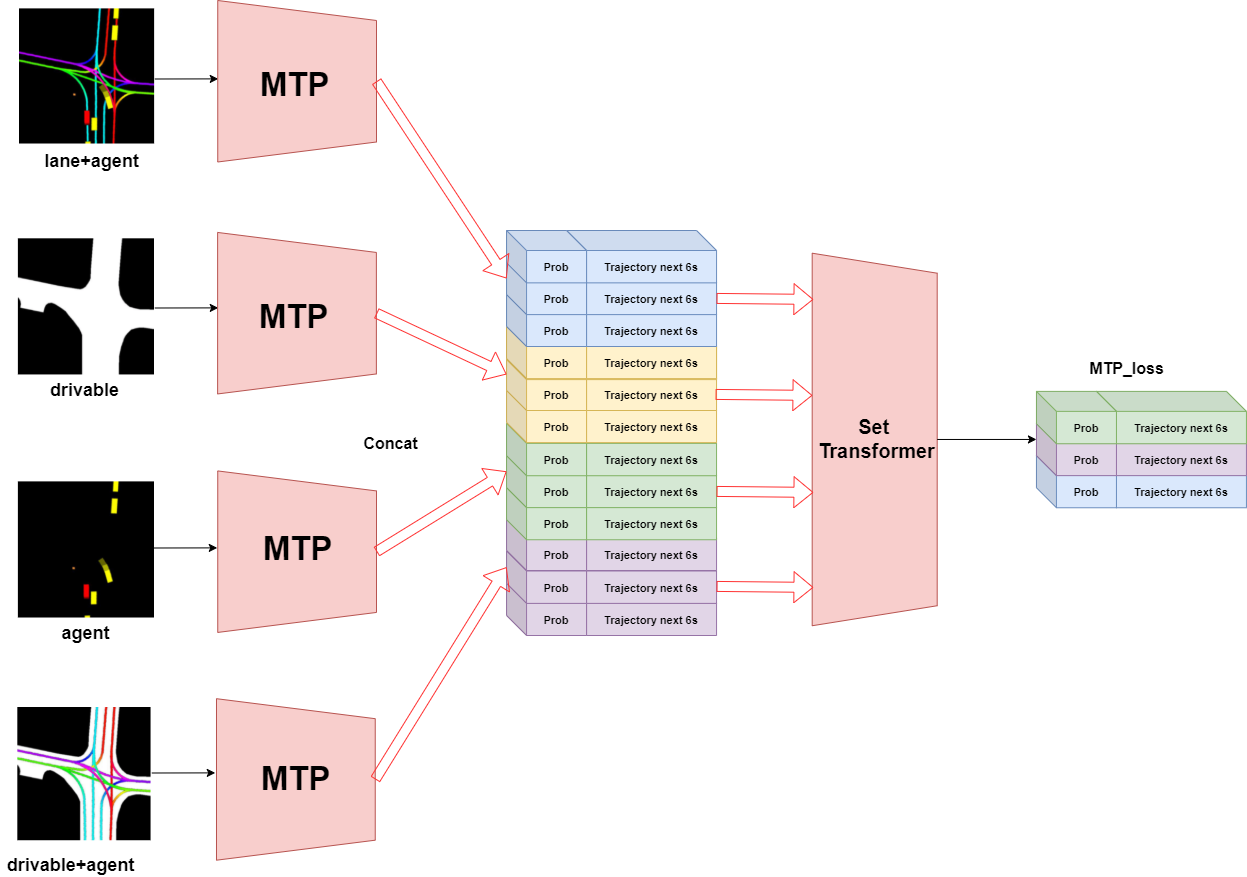}
\end{center}
  \caption{The architecture of ParallelNet.}
\label{fig:long}
\label{fig:onecol}
\end{figure*}

\textbf{Novel trajectory prediction methods:}
Rather than directly generating predicted trajectories and confidence, HOME ~\cite{gilles2021home} encode HD map by convolution layer as context encoding and ego and other agents history kinematic data as social encoding, then passing through upsample decod- ing layer, model outputs a heat map, depicting final desti- nation probability distribution. Then sample some destina- tions by finding optimal missing rate and final displacement error trade-off, finally compute trajectories for the selected destination. Our method follows the traditional way by di- rectly generating predicted trajectories and confidences, if not restricted time limit, we would attempt a probabilistic sampling approach to increase the trajectories variety.

\section{Method}
In this challenge, we use the 6h Nuscene prediction dataset ~\cite{caesar2020nuscenes} for multi-mode trajectory prediction, Nuscene dev-kit python library is provided, so that rasterized HD map with respect to any one of agent and its user-specified time step can be created and stored, its kinematic data, such as velocity, acceleration and head-turning rate can be queried. Therefore, we use rasterized HD map and its kine- matic data at the exact moment as input, then feed them as input to our neural network model. To train the model, a single ground truth trajectory is provided to supervise the training phase. Our work is following and extending the MTP ~\cite{cui2019multimodal} model.
\subsection{Model}
In our work, we first build the basic multi-trajectories prediction (MTP) model, as shown in Figure 1. We feed the rasterized HD map with all static and dynamic layers stacked together into its CNN backbone. Then pass through two fully connected layers for dimension projection, we obtain multiple predicted trajectories with confidences. However, we found that almost every predicted trajectory in the test set are straight ahead, for vehicle turning cases, trajectories can frequently point to undrivable areas, or even out of the map. Two reasons are causing this problem, first is the imbalanced type of scene and trajectory in the dataset. In most scene, the agent is moving straight ahead and following a straight lane, only in very few scenes, we could see the agent turning left, turning right, and turning head. So the model simply predicts straight head trajectories can cover most cases already and end training with a low validation loss. The second reason is due to information loss, since stacking multiple layers can cause overlap by dynamic layers on the top of some static layers, some important geometric features such as lanes can be lost or corrupted by agents on top of that, so predicted trajectories fail to identify geometric structure in such highly noisy HD maps, failing to predict curved trajectories to follow lanes and turn left or turn right.

To avoid feature extraction from noisy HD maps, we pro- posed ParallelNet in figure 2. to better make use of the ras- terized layers. There are 5 different types of layers, namely, drivable, lane, pedestrian crossing, walkway, agents. We rasterized all 5 types of layers, besides, we also rasterized some composite layers, drivable+lane, lane+agents, driv- able+agents, pedestrian crossing+walkway. Layer driv- able+lane can represent static layers that add constraints to the direction and curvature of predicted trajectories, layer lane+agents can represent the lane following principle for on-road agents, layer drivable+agents is similar, represent- ing the drivable area as a bound for on-road agents. Layer pedestrian crossing + walkway represents where on-road agents should slow down or stop, which acts as a security control for predicted trajectories. Finally, we trained sep- arate MTP models by using one single layer or one com- posite layer as input to analyze their guiding effect for pre-
dicted trajectories. As shown in Figure 3, in the trained MTP model, agents can roughly follow lanes on composite layers, thus we can see lane is an important feature for tra- jectory prediction in HD map. Also, to represent interaction relationship between agent and lane, we choose agents and agents+lane as part of our selected features. To allow our model to learn lane following and lane changing actions we select drivable+lane and lane as input features. We feed se- lected features into 4 parallel CNN backbone MobileNet v2~\cite{sandler2018mobilenetv2} for feature extraction, and each extracted feature from CNN backbone output is concatenated with kinematic data, then all four features pass through two fully connected lay- ers for dimension projection, we can obtain the predicted trajectories with confidences by each extracted feature, as ’independent hypothesis’ from four MTP models. Finally, we use Set Transformer ~\cite{lee2019set} for permutation invariant self- attention within all independent hypotheses, as a way of tra- jectories fusion and output the desired number of trajecto- ries.

\begin{figure}[h]
\begin{center}
\includegraphics[width=0.45\textwidth]{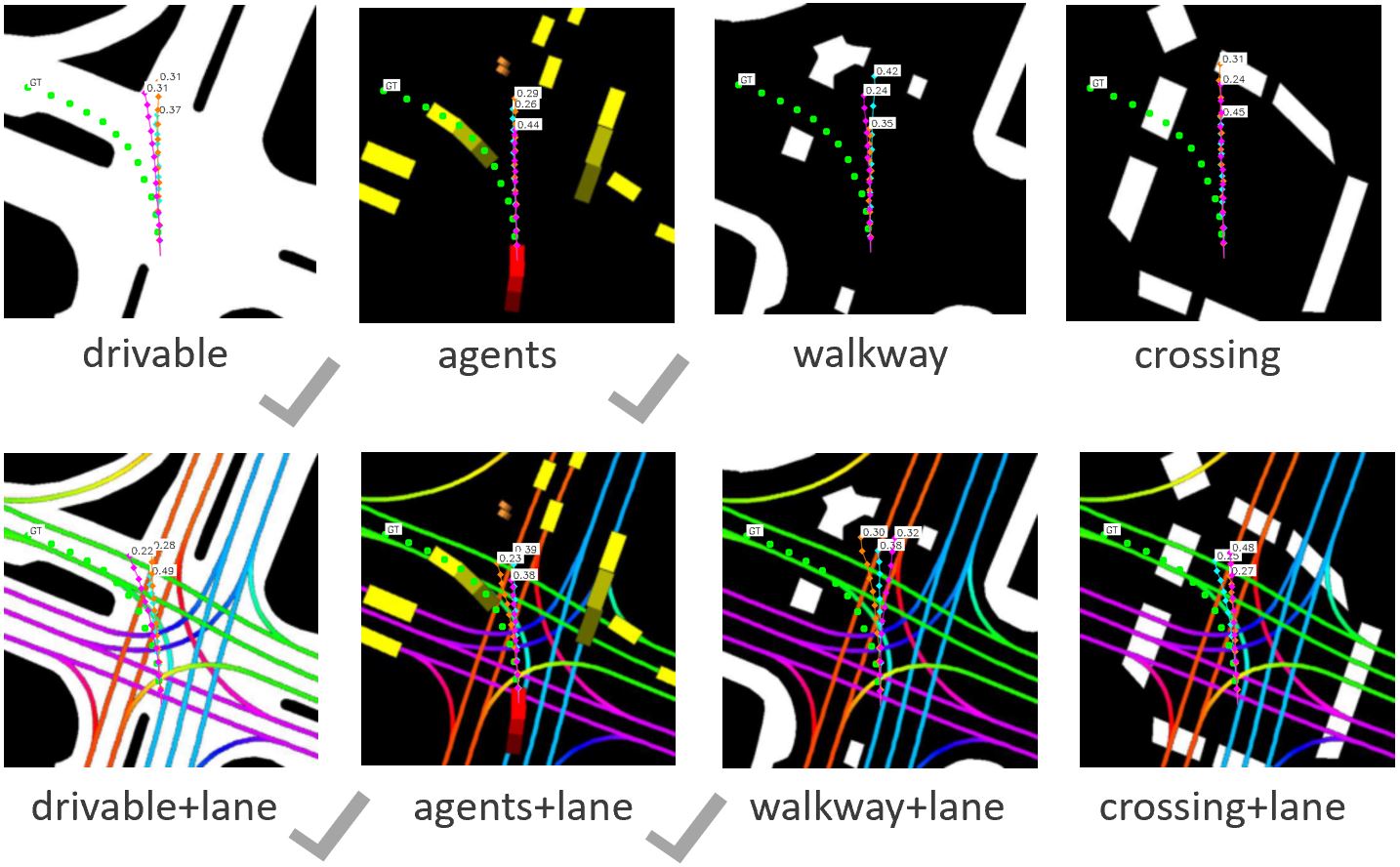}
\end{center}
  \caption{layer feature selection}
\label{fig:long}
\label{fig:onecol}
\end{figure}

\subsection{loss function}
We are predicting multiple trajectories with confidences, however, there is only one ground truth trajectory. We follow the loss function used in MTP model: given we have totally K trajectories and confidences, the future time length for each trajectory is T. the kth trajectory and confidence are expressed as follows:\\

$\displaystyle\text{Traj}_{\, k} = (x_1, y_1)_k , (x_2, y_2)_k, \cdots, (x_T, y_T)_k$ \\

$\displaystyle\text{Prob}_{\, k} = p_k, \;\;\sum^K_{k=1} p_k= 1.$\\

To use the ground truth, we should select the best predicted trajectory from the entire trajectory set and maximize its confidence. And the best trajectory is defined as the trajectory having least angle difference with ground truth:\\

$\displaystyle Traj^* = \underset{k\in\{1,\cdots,K\}}{\mathrm{argmin}}\{\theta(Traj_1, gt), \cdots,\theta(Traj_K, gt)\}$\\

$\displaystyle \theta(Traj_k, gt) = \cosh\Big(\frac{<(x_T, y_T)_k, ({x_T},{y_T})_{gt})>}{\|(x_T, y_T)_k\|_2*\|({x_T}, {y_T})_{gt}\|_2}\Big)$\\

Then we use MSE loss to regress the chosen best trajectory with ground truth, and maximize its confidence:\\

$\displaystyle\textit{Regression loss} = \frac{1}{T}\sum^T_{t=1}\|(x_t, y_t)-(\hat{x_t^*},\hat{y_t^*})\|_2$\\

$\displaystyle\textit{Classification loss} = -\log(p^*)$\\

$\displaystyle\textit{MTP loss} = \textit{Regression loss} + \textit{Classification loss}$\\

In addition, we introduce Angle scaled loss, to deal with dataset scene imbalance problem, larger steering angle is less frequent in dataset, thus we give it larger penalty. In \textbf{Figure 3}, it is clear that majority steering angles are close to 0, penalty will approach to 1, this case will barely add any additional loss. For minor cases where steering angle are larger, the frequency gets lower, thus penalty will exponentially increase. The angle scaled loss acts as a dataset sample imbalance inverse calibration, more infrequent a type of sample is, more penalty it will be given.\\

$\displaystyle\textit{Angle scaled loss = MTP loss}*\exp\Big(\frac{\alpha_{gt}}{20}\Big)$\\

$\displaystyle\alpha_{gt} = \tanh\Big(\frac{y_T}{x_T}\Big)$

\begin{table*}[]
\begin{tabular}{|c|c|c|c|c|c|c|}
\hline
Method                      & minADE5       & minADE10      & MissRateTop5  & MissRateTop10 & minFDE1       & offRoadRate \\ \hline
Angle scaled loss 12 modes  & \textbf{2.72} & \textbf{1.94} & \textbf{0.71} & \textbf{0.59} & \textbf{9.23} & 0.08        \\ 
MTP   loss 12 modes         & 3.29          & 2.09          & 0.77          & 0.62          & 10.7          & 0.08        \\ \hline
Angle   scaled loss 3 modes & 2.80          & 2.80          & 0.82          & 0.82          & 10.20         & 0.08        \\
MTP   loss 3 modes          & 2.86          & 2.86          & 0.82          & 0.82          & \textbf{9.23} & 0.08        \\ \hline
\end{tabular}
\caption{loss function ablation study}
\end{table*}

\begin{table*}[]
\centering
\begin{tabular}{|c|c|c|c|c|c|c|}
\hline
Method     & minADE5       & minADE10      & MissRateTop5  & MissRateTop10 & minFDE1       & offRoadRate   \\ \hline
2, 3,   4  & 3.29          & 2.09          & 0.77          & 0.63          & 10.76         & 0.09          \\ \hline
1, 2,   4  & 2.55          & \textbf{1.76} & 0.73          & \textbf{0.59} & 10.05         & \textbf{0.08} \\ \hline
1, 2,   3  & 2.49          & \textbf{1.76} & 0.71          & 0.62          & 9.88          & \textbf{0.08} \\ \hline
1, 2, 3, 4 & \textbf{2.36} & \textbf{1.76} & \textbf{0.67} & 0.60          & \textbf{9.82} & \textbf{0.08} \\ \hline
\end{tabular}
\caption{rasterized layer ablation study, 1:drivable+lane, 2:agents+lane, 3:drivable, 4:agents}
\end{table*}

\section{Experiments Backgrounds}
\subsection{Dataset}
We used the Nuscene prediction dataset, which contains 6 hours of driving data, including agent trajectories and corresponding HD maps for various types of 1000 scenes. Each scene consists of 20 seconds, and keyframes are collected at 2-hertz frequency. In each frame, objects are manually annotated, including position, size, class and visibility. Therefore, kinematic data of on-road agents can be computed and plays an important role in trajectory prediction. In our experiment, a single sample composes the rasterized HD map centred on a specific road agent, and its kinematic data, including velocity, linear acceleration and head-turning velocity, coupled with a ground truth trajectory in the next 6 seconds in 2-hertz frequency.\\

\subsection{Implementation Details}
We first use the Nuscene dev-kit to rasterize desired single or composite layers, and all of these layers are 300x300 and 3 channels. The optimizer is Adam, the learning rate is 1e-4, the batch size is 100, no fine-tune on the model. We train the model on Google Colab platform with a single NVIDIA V100 GPU, with 32GB VRAM, taking 1h to finish 20 epochs to converge. The model with best validation loss is stored and reloaded to make inference.\\
\subsection{Metrics}
Our model performance is evaluated on Nuscene prediction challenge server, there are totally 4 evaluation metrics:\\

\textbf{Minimum Average Displacement Error over k}. The average of pointwise L2 distances between the predicted trajectory and ground truth over the k most likely predictions, ADE follows formula below:\\

\textbf{ADE} = $\displaystyle\frac{1}{T}\sum^{T}_{t=1}\|(\hat{x_t}, \hat{y_t})-(x_t, y_t)\|_2$\\

\textbf{Minimum Final Displacement Error over k}.
The final displacement error (FDE) is the $L_2$ distance between the final points of the prediction and ground truth. We take the minimum FDE over the k most likely predictions and average over all agents. FDE follows formula below:\\

\textbf{FDE} = $\displaystyle\frac{1}{T}\|(\hat{x_T}, \hat{y_T})-(x_T, y_T)\|_2$\\

\textbf{Miss Rate At 2 meters over k}. 
If the maximum pointwise L2 distance between the prediction and ground truth is greater than 2 meters, we define the prediction as a miss. For each agent, we take the k most likely predictions and evaluate if any are misses, finally calculate the proportion of misses over all agents. The way to define a miss follows formula below:\\

\textbf{miss} = $\displaystyle \underset{t\in\{1,\cdots,T\}}{\mathrm{max}}\{\|(\hat{x_t}, \hat{y_t})-(x_t, y_t)\|_2\}>2.0$\\

\textbf{Off road rate}. The off road rate is the average off road frequency among all predictions, off road frequency for a single prediction is the number of trajectories shooting out of the road over the total number of predicted trajectories. \\

\section{Experiments Results}

\begin{table*}[]
\centering
\begin{tabular}{|c|c|c|c|c|c|c|}
\hline
Method             & minADE5       & minADE10      & MissRateTop5  & MissRateTop10 & minFDE1       & offRoadRate   \\ \hline
MTP                & 2.93          & 2.93          & 0.90          & 0.82          & \textbf{9.23} & 0.11          \\ 
CoverNet           & 2.62          & 1.92          & 0.76          & 0.64          & 10.03         & 0.13          \\ 
UKF Physics Oracle & 3.70          & 3.70          & 0.88          & 0.88          & 9.09          & 0.12          \\ 
ParallelNet        & \textbf{2.36} & \textbf{1.76} & \textbf{0.67} & \textbf{0.60} & 9.82          & \textbf{0.08} \\ \hline
\end{tabular}
\caption{ Compare to baseline methods, results are from Nuscene prediction challenge leaderboard.}
\end{table*}

\begin{table*}[]
\centering
\begin{tabular}{|c|c|c|c|c|c|c|}
\hline
Method          & minADE5       & minADE10      & MissRateTop5  & MissRateTop10 & minFDE1       & offRoadRate   \\ \hline
PGP             & \textbf{1.30} & \textbf{0.98} & \textbf{0.57} & \textbf{0.37} & 7.72          & \textbf{0.03} \\ 
HOME            & 1.42          & 1.15          & \textbf{0.57} & 0.47          & \textbf{6.99} & 0.04          \\ 
UCSD-LISA (P2T) & 1.45          & 1.16          & 0.64          & 0.46          & 10.50         & \textbf{0.03} \\ 
ParallelNet     & 2.36          & 1.76          & 0.67          & 0.60          & \textbf{9.82} & 0.08          \\ \hline
\end{tabular}
\caption{Compare to the most recent state of art model, results are from Nuscene prediction challenge leaderboard.}
\end{table*}
\subsection{Ablation Study}
We conducted a series of experiments to validate our de- sign choices. The first is to compare our newly introduced loss function with previous MTP loss by both sparse (3 modes) and dense (12 modes) trajectory prediction. As the result of Table 1 shows, angle scaled loss can not showcase an advantage in sparse prediction over MTP loss. However,in dense trajectory prediction, it is clear that almost all met- rics angle scaled loss outperform MTP loss. The reason is that increase from 3 modes to 12 modes increase the poten- tial variety of predicted trajectories. When only 3 modes are predicted, there could hardly be any variety and both mod- els are inclined to predict straight lines. However, when increase to 12 modes, as we can see in Figure 4, Angle scaled loss model tend to predict trajectories with differ- ent turning angles especially at the crossroad, but MTP loss fails to increase as much variety as Angle scaled loss model, thus significantly increase the prediction hit rate on ground truth. Therefore, for minADE, minFDE and missRate, it is reasonable to see better results for our Angle scaled loss model.\\
\begin{figure}[h]
\begin{center}

\includegraphics[width=0.4\textwidth]{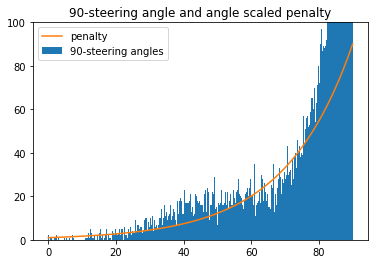}
\end{center}
  \caption{Angle scaled loss}
\label{fig:long}
\label{fig:onecol}
\end{figure}
Our second ablation test focus on our previously cho- sen rasterized layers, we want to study the effect of each layer by adding or removing it. From Table 2, unexpectedly when using all four layers, the model has the best perfor- mance, because we ensemble four models rather than three. The most notable thing is, unlike the other two 3 model ensembles, we observe there is a great performance drop with- out using layer drivable+lane, we can see that in almost all metrics the score gets much poorer. The reason is clear since drivable+lane is a static layer, which does not change by varying timestamp in our dataset. The information kept by this layer can help the model learn the general pattern of driving: either keep following a lane or cross the drivable area and follow another lane. With the loss of this static layer, even though lane+agent contains information about the lane, but it is no longer static, the rasterized agents can cover lots of regions of the lane, which makes the lane fea- ture highly noisy and this prevent the model to learn lane following or lane changing effectively. So we can conclude that drivable+lane is the most important feature in all four chosen layers.\\

\subsection{Results}
We compare our work to three baseline models in the
Nuscene prediction challenge: 1. MTP model, published on
ICRA 2019, is the basis that our work built upon. 2. Physics Oracle, a completely kinematics based method without any semantic information such as HD maps, has been highly optimized and trained by vast amount of driving data. 3. CoverNet~\cite{phan2020covernet}, one of the state of art models for multimode trajectory prediction by greedy selection from generated trajectory set, published on CVPR 2020. As the result shown on Table 2, our model significantly outperforms all the baselines. The main reason is probably that none of these models separates rasterized layers, and their feature extraction is lossy and noisy, However, baseline models have much lighter network architecture and thus faster inference time, but it does not mean our model is impractical for real-time inference. Based on our test, our model can make inference in average 5.4 ms for a single scene parsing on an NVIDIA V100 GPU with batch size =100. Also, our model does not use history rasterized scenes, therefore our input is lighter than CoverNet. Moreover, our model is parallelizable during the scene parsing phase, forward pass can be synchronized on all four backbones by adapting multiple GPU, to significantly reduce computational time overhead.
\begin{figure}[h]
\begin{center}

\includegraphics[scale=0.45]{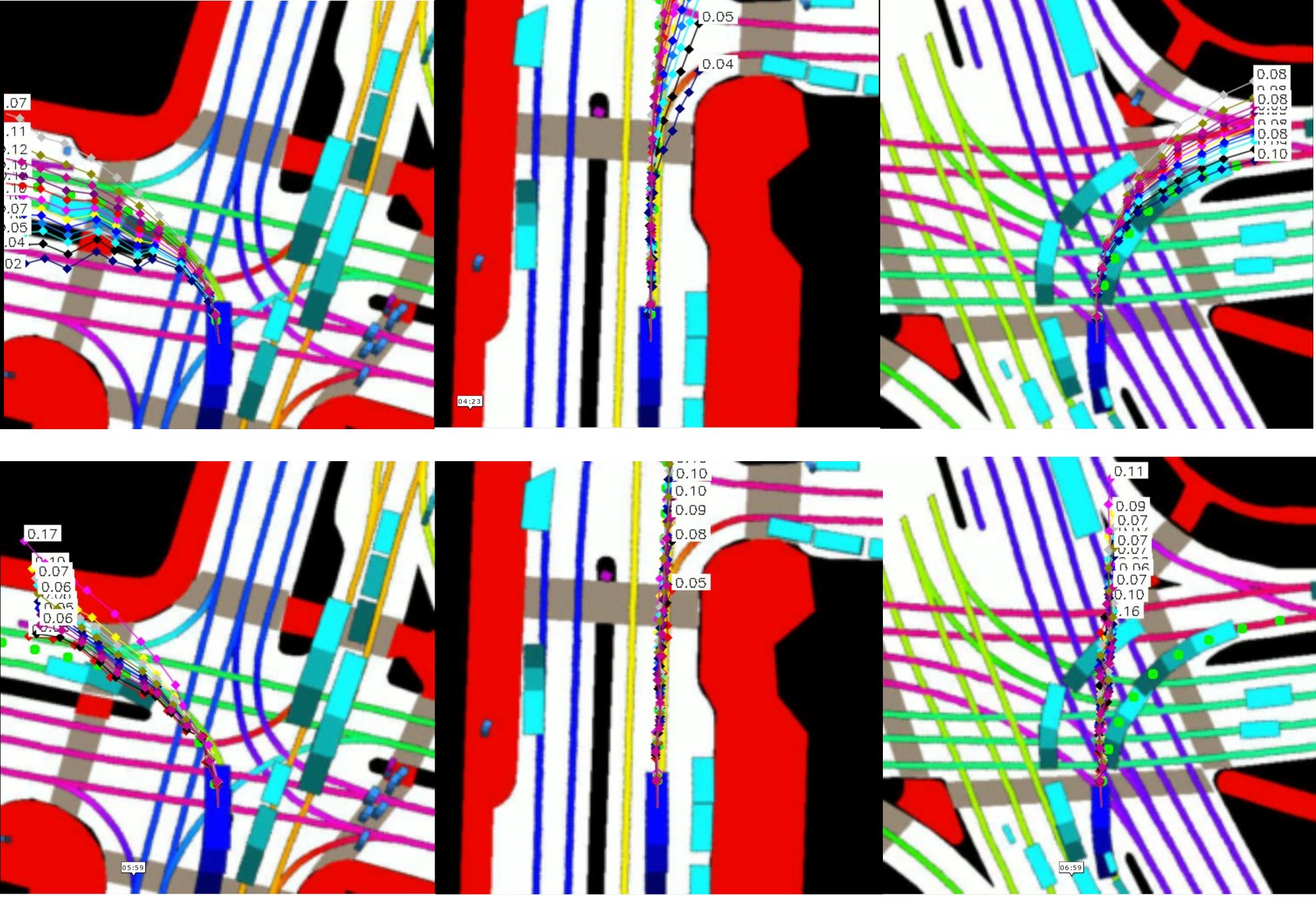}
\end{center}
  \caption{12 mode: Angle scaled loss row 1, MTP loss row 2.}
\label{fig:long}
\label{fig:onecol}
\end{figure}

\begin{figure*}[h]
\begin{center}

\includegraphics[width=\textwidth,,height=9cm]{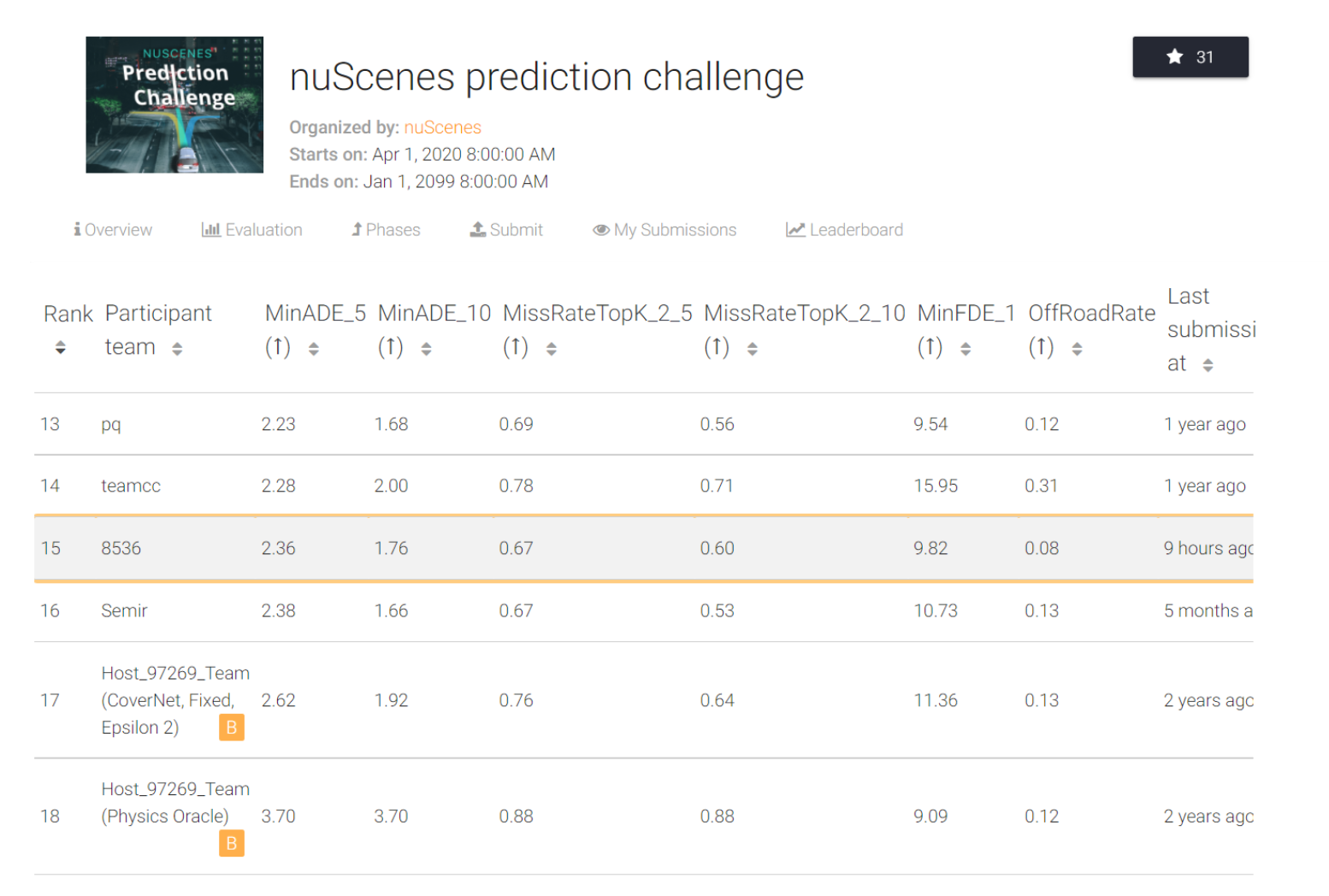}
\end{center}
\label{fig:long}
\label{fig:onecol}
\caption{The NuScene 2020 Prediction Challenge Leaderboard.}
\end{figure*}

Besides, we also submitted our prediction results to Nuscenes Prediction Challenge 2020, and our latest ranking is 15th, which is better than two baseline model CoverNet and UKF. See Figure 6 for more details.

In Figure.7, we visualize our predication performance, and there are 8 samples picked from test set. As mentioned before, our model can provide 3 trajectory predictions, compared with GT trajectory, our prediction always be correct when facing with straight line and corner road. But based on current vehicle position, when there is ambiguity and are multiple possibility to turn left or go straight like first visualization sample at first row in Fig.7 which the GT is going to left, but our model predicate to going straight.    

Finally, we had an overview of the most recent state of the art model evaluated on the Nuscene dataset. PGP ~\cite{deo2022multimodal} models the inherent uncertainty in driving behaviour, and decouple it as lateral uncertainty (keep lane, turning) and longitudinal uncertainty (accelerating, braking). The model combines learned driving policy, capturing the lateral uncertainty and decoded lane graph information, capturing the longitudinal uncertainty to generate multi-mode trajectory prediction. HOME ~\cite{gilles2021home} as shown in related work section, generate Heat Map to represent trajectories distribution, then sample from the Heat Map for optimal missing rate and final displacement error trade-off. P2T ~\cite{deo2020trajectory} models multimode trajectory prediction as a sequential decision making problem, using maximum entropy inverse reinforcement learning to learn a grid-based sampling method for trajectory planning. From those most recent advancements, we can see some trends: Rasterized maps are no longer directly fed into neural network models, lane graph representation is replacing the semantic map. Explicit driving behaviour categorization and modelling become more popular since it explicitly captures the driving behaviour diversity and uncertainty. More and more engineering flavours: instead of relying more on neural networks for feature extraction, human knowledge in driving and traffic modelling is more efficient in accompanying the neural network model. Our model completely relies on neural network model automatic feature extraction, therefore our model performance is far away from the most recent state of the art.

\begin{figure*}[h]
\begin{center}

\includegraphics[width=\textwidth,height=7cm]{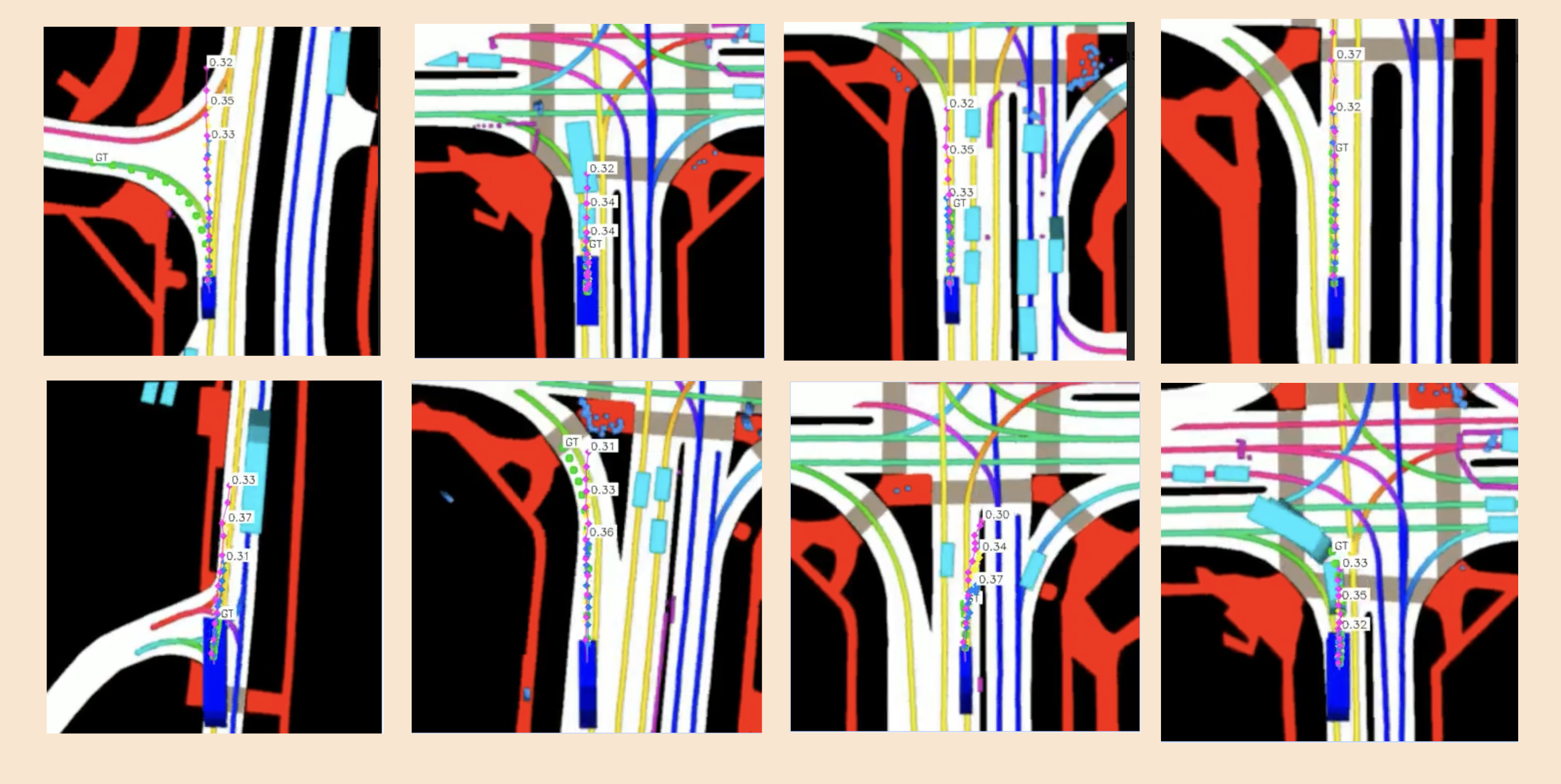}
\end{center}
\caption{Visualize Performance}
\label{fig:long}
\label{fig:onecol}
\end{figure*}

\section{Conclusion}
We proposed new network architecture for multi-mode trajectory prediction accompanied by a new loss function Angle scaled loss. Our model feeds multiple types of single or composite rasterized layers into separate MTP model, then fuse all predicted trajectories and their corresponding confidences by Set Transformer reduced to the desired number of modes. We conducted two ablation studies on our newly introduced loss and demonstrate its effectiveness in predicting vehicle steering by comparing it with MTP loss, and models with different input features, and found the necessity of using the feature from the static layer. Finally, we compare our work with baseline models listed on the Nuscene prediction challenge leader board, our model easily beats the MTP model, UKF Physics Oracle model, and CoverNet with less input data, with computational overhead as a trade-off\\

Our Future work will focus on replacing the parallel backbone structure, a low-efficiency feature extraction process with explicit lane modelling without the necessity of feature extraction. Also, attempt different agent modelling methods, for example, split with agents in the environment and focus on ego agent’s behaviour modelling, or treat all on-road objects as a multi-agent system, learn and predict the joint position distribution for future seconds.

{\small
\bibliographystyle{ieee_fullname}
\bibliography{egbib}

\begin{thebibliography}{10}\itemsep=-1pt

\bibitem{caesar2020nuscenes}
Holger Caesar, Varun Bankiti, Alex~H Lang, Sourabh Vora, Venice~Erin Liong,
  Qiang Xu, Anush Krishnan, Yu Pan, Giancarlo Baldan, and Oscar Beijbom.
\newblock nuscenes: A multimodal dataset for autonomous driving.
\newblock In {\em Proceedings of the IEEE/CVF conference on computer vision and
  pattern recognition}, pages 11621--11631, 2020.

\bibitem{cui2021lookout}
Alexander Cui, Sergio Casas, Abbas Sadat, Renjie Liao, and Raquel Urtasun.
\newblock Lookout: Diverse multi-future prediction and planning for
  self-driving.
\newblock In {\em Proceedings of the IEEE/CVF International Conference on
  Computer Vision}, pages 16107--16116, 2021.

\bibitem{cui2019multimodal}
Henggang Cui, Vladan Radosavljevic, Fang-Chieh Chou, Tsung-Han Lin, Thi Nguyen,
  Tzu-Kuo Huang, Jeff Schneider, and Nemanja Djuric.
\newblock Multimodal trajectory predictions for autonomous driving using deep
  convolutional networks.
\newblock In {\em 2019 International Conference on Robotics and Automation
  (ICRA)}, pages 2090--2096. IEEE, 2019.

\bibitem{deo2020trajectory}
Nachiket Deo and Mohan~M Trivedi.
\newblock Trajectory forecasts in unknown environments conditioned on
  grid-based plans.
\newblock {\em arXiv preprint arXiv:2001.00735}, 2020.

\bibitem{deo2022multimodal}
Nachiket Deo, Eric Wolff, and Oscar Beijbom.
\newblock Multimodal trajectory prediction conditioned on lane-graph
  traversals.
\newblock In {\em Conference on Robot Learning}, pages 203--212. PMLR, 2022.

\bibitem{gao2020vectornet}
Jiyang Gao, Chen Sun, Hang Zhao, Yi Shen, Dragomir Anguelov, Congcong Li, and
  Cordelia Schmid.
\newblock Vectornet: Encoding hd maps and agent dynamics from vectorized
  representation.
\newblock In {\em Proceedings of the IEEE/CVF Conference on Computer Vision and
  Pattern Recognition}, pages 11525--11533, 2020.

\bibitem{gilles2021home}
Thomas Gilles, Stefano Sabatini, Dzmitry Tsishkou, Bogdan Stanciulescu, and
  Fabien Moutarde.
\newblock Home: Heatmap output for future motion estimation.
\newblock In {\em 2021 IEEE International Intelligent Transportation Systems
  Conference (ITSC)}, pages 500--507. IEEE, 2021.

\bibitem{lee2019set}
Juho Lee, Yoonho Lee, Jungtaek Kim, Adam Kosiorek, Seungjin Choi, and Yee~Whye
  Teh.
\newblock Set transformer: A framework for attention-based
  permutation-invariant neural networks.
\newblock In {\em International conference on machine learning}, pages
  3744--3753. PMLR, 2019.

\bibitem{liang2020learning}
Ming Liang, Bin Yang, Rui Hu, Yun Chen, Renjie Liao, Song Feng, and Raquel
  Urtasun.
\newblock Learning lane graph representations for motion forecasting.
\newblock In {\em European Conference on Computer Vision}, pages 541--556.
  Springer, 2020.

\bibitem{ngiam2021scene}
Jiquan Ngiam, Benjamin Caine, Vijay Vasudevan, Zhengdong Zhang, Hao-Tien~Lewis
  Chiang, Jeffrey Ling, Rebecca Roelofs, Alex Bewley, Chenxi Liu, Ashish
  Venugopal, et~al.
\newblock Scene transformer: A unified multi-task model for behavior prediction
  and planning.
\newblock {\em arXiv e-prints}, pages arXiv--2106, 2021.

\bibitem{phan2020covernet}
Tung Phan-Minh, Elena~Corina Grigore, Freddy~A Boulton, Oscar Beijbom, and
  Eric~M Wolff.
\newblock Covernet: Multimodal behavior prediction using trajectory sets.
\newblock In {\em Proceedings of the IEEE/CVF Conference on Computer Vision and
  Pattern Recognition}, pages 14074--14083, 2020.

\bibitem{qi2017pointnet}
Charles~R Qi, Hao Su, Kaichun Mo, and Leonidas~J Guibas.
\newblock Pointnet: Deep learning on point sets for 3d classification and
  segmentation.
\newblock In {\em Proceedings of the IEEE conference on computer vision and
  pattern recognition}, pages 652--660, 2017.

\bibitem{sandler2018mobilenetv2}
Mark Sandler, Andrew Howard, Menglong Zhu, Andrey Zhmoginov, and Liang-Chieh
  Chen.
\newblock Mobilenetv2: Inverted residuals and linear bottlenecks.
\newblock In {\em Proceedings of the IEEE conference on computer vision and
  pattern recognition}, pages 4510--4520, 2018.

\bibitem{zhao2021tnt}
Hang Zhao, Jiyang Gao, Tian Lan, Chen Sun, Ben Sapp, Balakrishnan Varadarajan,
  Yue Shen, Yi Shen, Yuning Chai, Cordelia Schmid, et~al.
\newblock Tnt: Target-driven trajectory prediction.
\newblock In {\em Conference on Robot Learning}, pages 895--904. PMLR, 2021.

\end{thebibliography}
}

\end{document}